# Prototipo de un Contador Bidireccional Automático de Personas basado en sensores de visión 3D

Benjamín, Ojeda-Magaña; Rubén, Ruelas; José G, Robledo-Hernández; Víctor M, Rangel Cobián; Fernando, López Aguilar-Hernández

Universidad de Guadalajara, Departamento de Ingeniería de Proyectos, CUCEI

**Resumen**

Los sensores 3D, de profundidad o también conocidos como RGB-D basan su funcionamiento en una imagen o matriz de profundidad, en donde cada pixel determina la distancia que hay entre la cámara 3D y los objetos, a partir de diferentes principios como: luz estructurada, tiempo de vuelo, estereovisión, entre otros. Hoy en día, debido a los avances en visión artificial desarrollados en los últimos años, existen cámaras 3D de bajo costo, las cuales permiten la captura de imágenes en "tiempo real" que son capaces de detectar un objeto, sin necesidad de que exista un movimiento en dicho objeto. Lo que nos permite extraer mayor y mejor información que con una cámara 2D. Por lo tanto, las cámaras 3D permiten detectar objetos de distintos colores y reflectividad dentro de una misma imagen, y evitar los problemas con los cambios de iluminación (en especial cuando se requiere detectar movimiento).

El objetivo del presente prototipo, consiste en realizar el conteo de personas de forma bidireccional (ingreso/salida) en un recinto a través de sensores de profundidad RGB-D. Este prototipo tiene aplicaciones especialmente para temas de seguridad y vigilancia en diferentes espacios, como: edificios, estadios, auditorios, centros comerciales, aeropuertos, salones de fiesta entre otros. La propuesta consiste en conocer en tiempo real el número de personas que se encuentran dentro de un recinto, y determinar si se sobrepasa el aforo máximo del lugar, lo cual es muy importante en caso de que sobrevenga un desastre como: un incendio, o un terremoto. Otra característica favorecedora del prototipo, consiste que, en el caso de eventos multitudinarios, se puede determinar si hay más personas de las que indica el boletaje o la capacidad del lugar, así como otras aplicaciones adicionales que se pueden desarrollar.

Por lo tanto; en este documento se describe el desarrollo de un prototipo para el conteo bidireccional automático de personas (entradas y salidas). El prototipo es un sistema portátil que está integrado por una cámara de profundidad *RealSense* D415, instalada de manera cenital en el macro de una puerta, para la captura de mapas de profundidad, de entrada/salida de un recinto. Además, se utiliza una mini computadora, que analiza la información ejecutando un algoritmo de detección de objetos mediante el movimiento y medición de la distancia a los mismos (personas) presentes en el campo de visión de la cámara. El algoritmo desarrollado para el prototipo permite el conteo bidireccional de personas, además permite almacenar la información para poder realizar análisis estadísticos, como por ejemplo determinar cuántas personas e encuentran dentro de un recinto. Adicionalmente, el prototipo también dispone con una cámara transversal 2D para que cuando se detecta que una persona atraviesa la región de interés, se capture una imagen con la cual se puede verificar quién o quiénes han entrado o salido del recinto. Para la implementación de los algoritmos se utilizaron lenguajes de programación de alto nivel como C++, Python y PHP, mientras que para la gestión de imágenes se utilizaron las bibliotecas de OpenCV más las bibliotecas propias del fabricante de Intel.

# 1. Introducción

Un contador de personas es un sistema que permite determinar el número de personas que ingresan o salen en una región determinada denominada zona de detección. En sus orígenes esta tarea se realizaba de forma manual, pero el desarrollo de la tecnología ha permitido utilizar diversos recursos mecánicos, electrónicos e informáticos para llevar a cabo dicha tarea de manera automática, ejemplos de lo anterior los tenemos en torniquetes, barras con sensores infrarrojos, básculas, cámaras 2D, etc. En este proyecto se presenta un contador de personas automático que utiliza técnicas de visión artificial por computadora para realizar dicho conteo, utilizando la secuencia de imágenes recogidas con un sensor de profundidad Intel®*RealSense* D415, así como un estudio sobre el número máximo de personas que debe de haber un determinado recinto (escuela, auditorio, salón de eventos, etc.)

El conteo bidireccional de personas (ingresan/salen) es una característica deseable para un sistema automático con aplicaciones potenciales en múltiples escenarios, por ilustrar con algunos ejemplos, es necesario conocer el número de pasajeros que entran y salen de un medio de transporte público (autobús, tren ligero, microbús, entre otros) para llevar a cabo su control y gestión. Por otro lado, en discotecas, edificios públicos, escuelas, hospitales, centros comerciales, entre otros, los protocolos de evacuación están diseñados de acuerdo con la capacidad del recinto, no debiendo sobrepasarse el número de personas máximo permitido, para evitar situaciones peligrosas, como por ejemplo un terremoto, en donde es necesario conocer el número de personas que están dentro del recinto para realizar obras de rescate, y que éstas no sean en vano. El control de presencia/ausencia al entrar una persona a una estancia, o sala, es también esencial para la implantación de políticas de ahorro energético. En otro contexto, la obtención automática de datos, del número de personas que hay en un recinto en tiempo real, así como la medida de tiempos de atención, como por ejemplo en bancos comerciales, es también de gran interés para empresarios y anunciantes.

Debido a las diversas aplicaciones y su creciente interés, en la literatura reciente se describen distintas soluciones para conocer, lo más exactamente posible, el número de personas que accede y sale de un espacio delimitado. Las dos principales tecnologías que los investigadores han utilizado hasta la fecha para resolver este problema son: a) imágenes de los escenarios analizadas con técnicas de visión por computadora y b) sensores que emiten un haz de luz IR (emisor/receptor), y que, al ser interrumpido, determina la entrada o salida de personas.

Los sistemas basados en emisión de haces de luz tienen la ventaja de preservar la privacidad de las personas, al no capturar imágenes. Este tipo de sistemas están integrados en los autobuses y locales comerciales, y son barras contadoras, que sólo permiten contar una sola persona a la vez. Otra manera de contar personas con haces de luz, es colocando unos sensores en el techo y otros en el muro, y de esta manera se pude determinar el número de personas que ingresan o salen, además de conocer su identidad.

Los sistemas basados en visión artificial, pueden aplicarse con menos restricciones en lugares más abiertos, como estacionamiento, centros comerciales, hospitales, edificios gubernamentales, entre otros. Dependiendo de la ubicación de la cámara, se distinguen entren sistemas cenitales, y no cenitales o frontales. El problema con los sistemas no cenitales es que no pueden ser utilizados en aplicaciones donde la privacidad sea un requerimiento a cumplir.

Al contrario, la configuración cenital es capaz, en principio de preservar la privacidad debido a que los rostros de las personas no son capturados.

## 2. Características e implementación del prototipo

En enero del 2018 Intel lanzo la cámara con sensor de profundidad D415 (ver Figura 1), que forma parte de la familia D400. A diferencia de otra cámara popular de profundidad que funciona en el controlador de juegos Kinect I y II, los modelos de Intel pueden funcionar con cualquier dispositivo, y están destinados principalmente a desarrolladores e ingenieros. Las cámaras de la familia D400, funcionan con USB III y constan de un par de sensores de profundidad, un sensor de RGB y un proyector infrarrojo. La cámara de profundidad Intel RealSense ha sido diseñada para equipar dispositivos con la capacidad de ver, comprender, interactuar y aprender de su entorno.

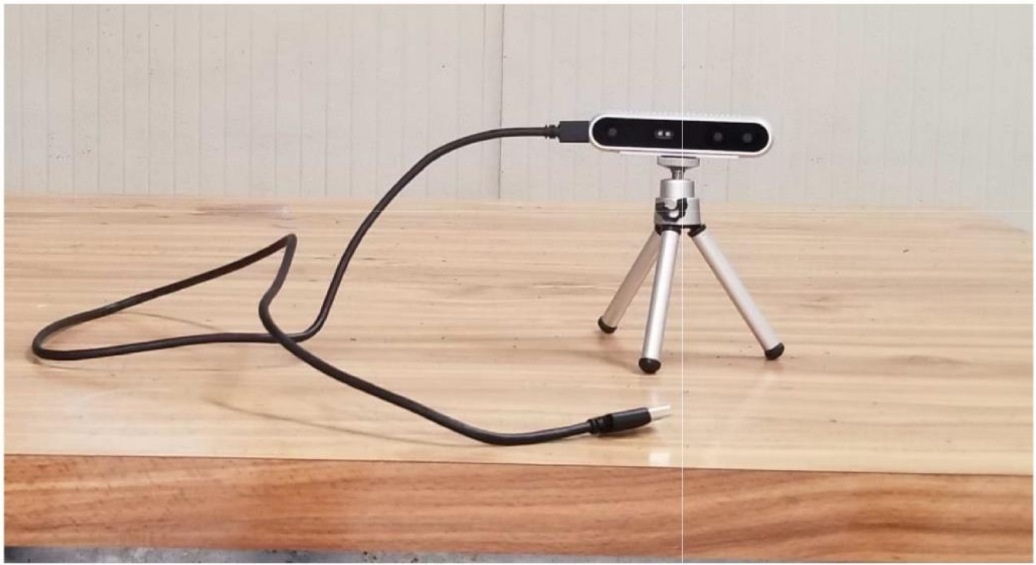

**Figura 1.** Sensor de profundidad Intel RealSense D415 utilizado en este prototipo

Las cámaras equipadas con sensores de profundidad infrarrojos obtienen imagen de los objetos teniendo en cuenta la profundidad a la que se encuentran los objetos, por lo tanto construyen una imagen tridimensional de la escena en el fotograma, desde un punto específico listas para usarse, y se pueden agregar fácilmente a los prototipos existentes vía USB, que incluyen visión de largo alcance, y funciona tanto, en áreas interiores como exteriores.

La tecnología de Intel RealSense admite una amplia variedad de sistemas operativos y lenguajes de programación. El Kit de Desarrollo de Software (SDK, por sus siglas en inglés) le permite extraer datos de profundidad de la cámara y utilizar la interpretación de estos en la plataforma de elección.

**Diseño del prototipo**

En la Figura 2 se puede observar un diagrama que muestra la ubicación y el posicionamiento de la cámara RGB-D la cual realizará el conteo de personas desde una posición cenital. Además, se observa una línea de color café, que es un umbral, que indica que la cámara RGB-D, sólo ve objetos a una distancia de 1 metro, y los objetos que sean mayores a un metro no los identifica.

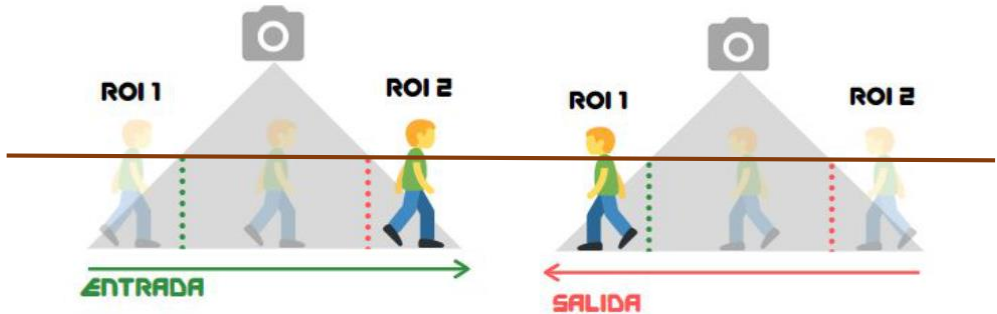

**Figura 2**. Diagrama vista lateral a) ingreso b) salida.

Las herramientas empleadas a nivel físico son las cámaras 2D y RGB-D. La cámara de Intel *RealSense* utilizada para el prototipo admite una amplia variedad de sistemas operativos y lenguajes de programación. Intel *RealSense* SDK 2.0 nos permite extraer los datos de la profundidad de la cámara e interpretarlos.

**Diagrama de flujo de Algoritmo de conteo de ROI´s (regiones de interés)**

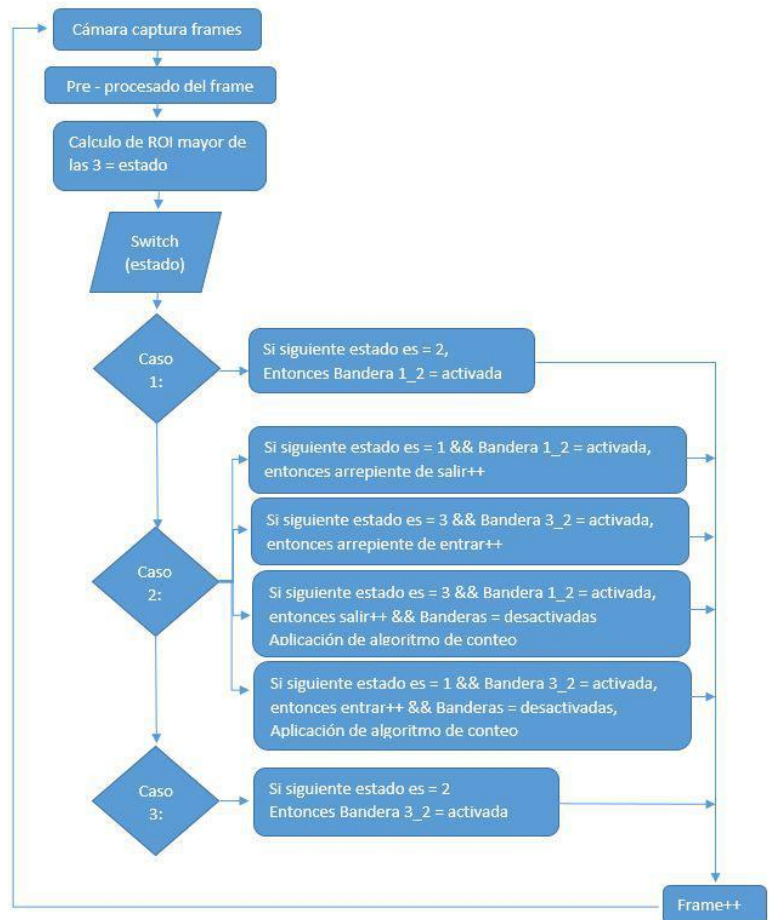

**Figura 3**. Diagrama de flujo del algoritmo en C++ del contador de personas.

Como se puede observar en la Figura 3, el procesamiento de las imágenes (*frames*) es iterativo, se realizan una serie de acciones para determinar en qué región de interés se encuentra el objeto en movimiento (persona) y a partir de esta información se sigue un patrón que como resultado arrojará si una persona entra, sale o se arrepiente de entrar o salir.

### 3. Pseudocódigo Algoritmo de Conteo de Personas

En la Figura 4 podemos observar el funcionamiento del algoritmo como pseudocódigo que conceptualiza de manera simple las acciones realizadas durante el procesamiento de imágenes para obtener un conteo bidireccional.

```
Cámara captura frames
Por cada frame se pre-procesa la imagen
    Convertir a escala de grises la imagen
Binarización de cada ROI 1,2,3
    Calculo de la ROI mayor es igual al estado
    Si estado es igual a 1
        Si siguiente estado es igual a 2,
            Entonces Bandera 1_2 es activada.
    Si estado es igual a 2
        Si siguiente estado es igual a 1 y bandera 1_2 es igual a activada,
            Entonces persona se arrepiente de salir ++
        Si siguiente estado es igual a 3 y bandera 3_2 es igual a activada,
            Entonces persona se arrepiente de entrar++
        Si siguiente estado es igual a 1 y bandera 3_2 es igual a activada,
            Entonces persona entra++
            Bandera 1_2 igual a 0 && Bandera 3_2 igual a 0.
                Persona sale – Persona entra es igual a Personas dentro
        Si siguiente estado es igual a 3 y Bandera 1_2 es igual a activada,
            Entonces Persona sale++
            Bandera 1_2 igual a 0 && Bandera 3_2 igual a 0.
                Persona sale – Persona entra es igual a Personas dentro.
    Si estado es igual a 3
        Si siguiente estado es igual a 2,
            Entonces Bandera 3_2 es activada.
Frame++
```

**Figura 4**. Pseudocódigo de Conteo de Personas.

**Diseño de interfaz web**

En la última fase de desarrollo, con el fin de proporcionar en tiempo real el conteo de personas, se realizó una interfaz web con HTML5 y PHP utilizando JavaScript y Ajax para ejecutar scripts de actualización de datos, lo que permite visualizar desde cualquier dispositivo móvil o fijo conectado a la Red de Internet, saber el número de personas que ingresaron o salieron del recinto.

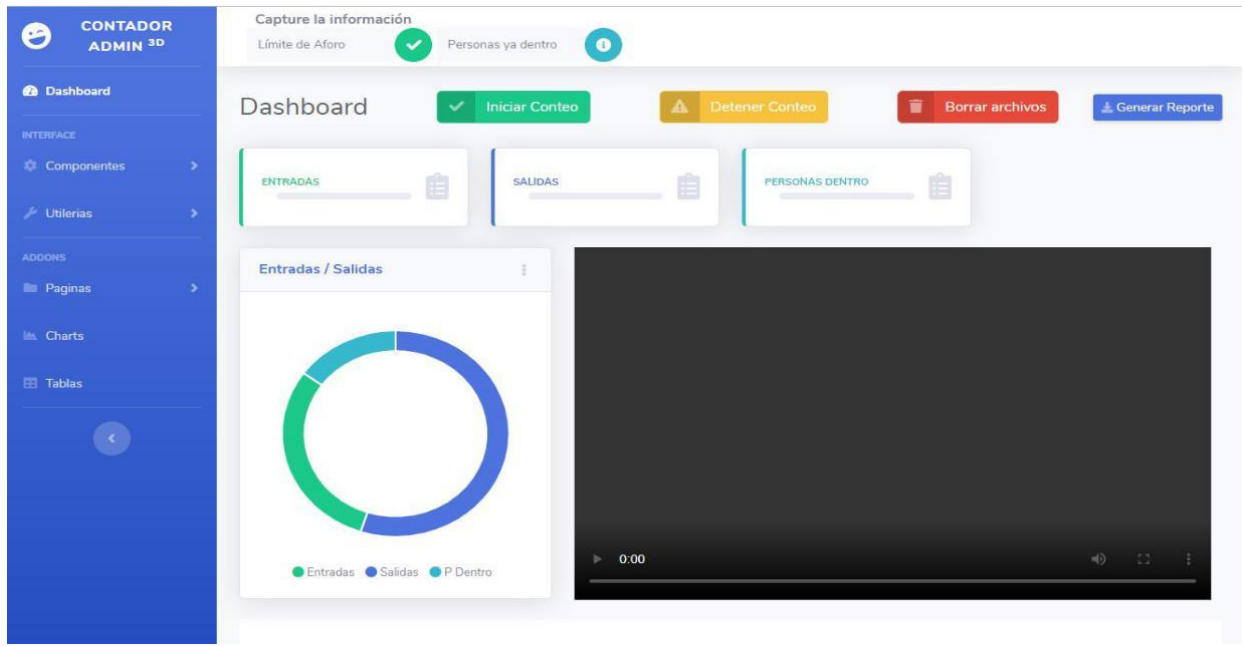

**Figura 5**. Interfaz Web de Contador de Personas.

En la Figura 5 se muestra la organización de los componentes del sistema contador de personas. Una venta que ofrece la interfaz web es: que es posible ejecutar el programa, detenerlo, borrar los archivos de entrada o salida generados, y generar reportes de análisis del procesamiento de imágenes.

En la Figura 6 se muestra con detalle los indicadores del sistema durante una prueba de conteo en un recinto.

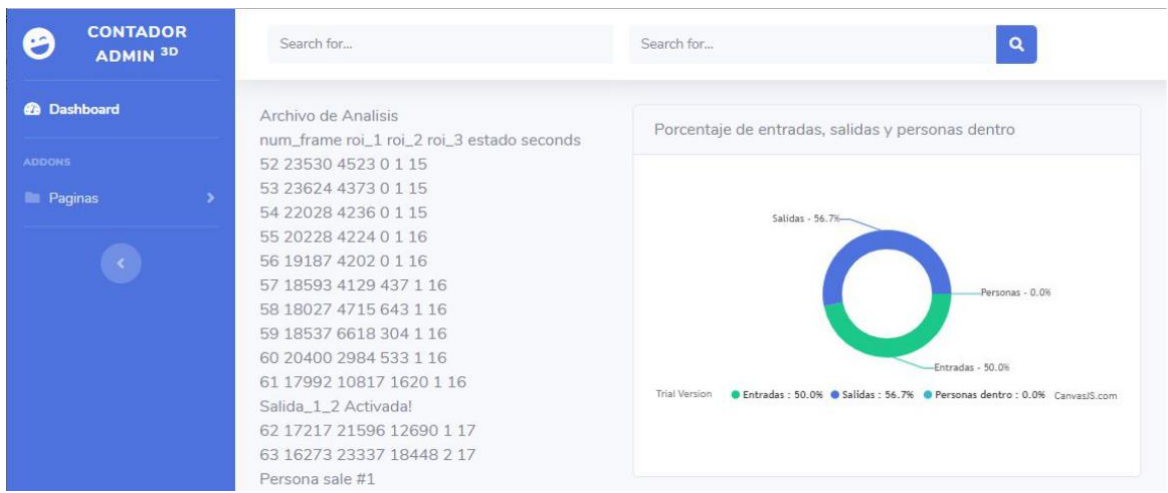

**Figura 6**. Visualización de archivo de análisis del conteo de personas y gráfica.

El archivo de análisis como se puede observar en la Figura 6, muestra las activaciones de las regiones de interés que detectó el algoritmo. Este archivo guarda los datos crudos del análisis y es posible verificar desde este los errores que pudieron haber ocurrido durante la ejecución del programa.

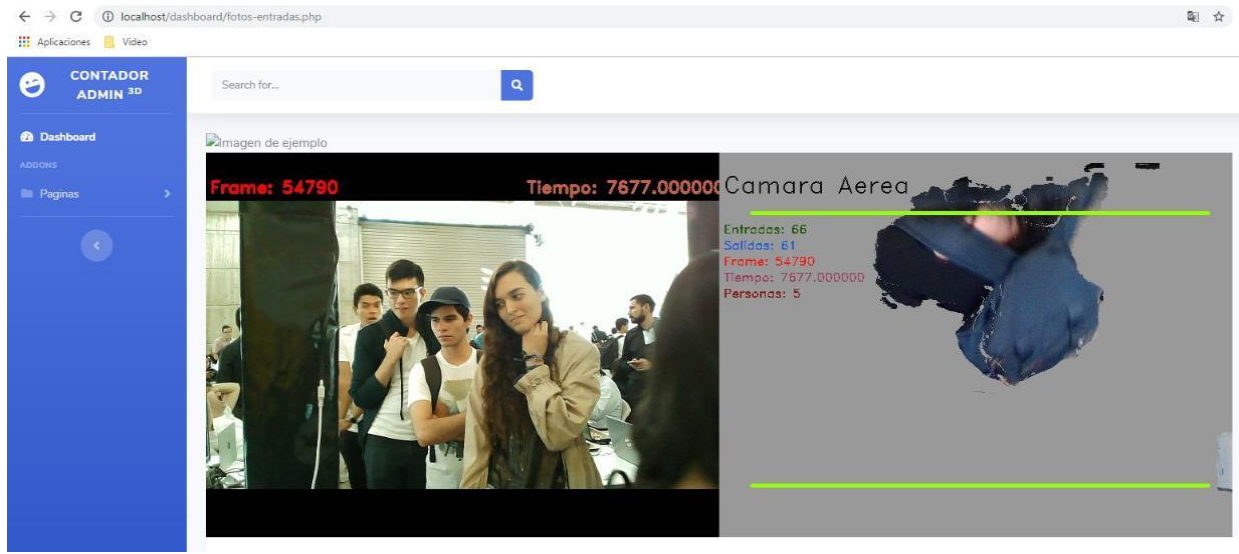
**Figura 7**. Interfaz web, mostrando la entrada de una persona.

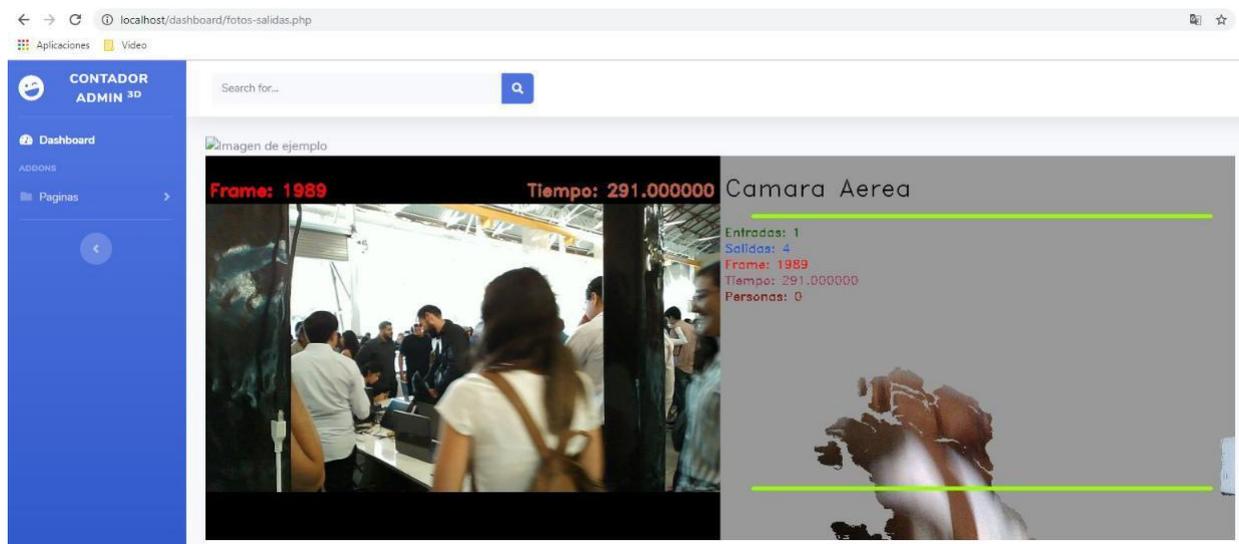
Figura 8. Interfaz web, mostrando la salida de una persona.

Las Figuras 7 y 8 muestran dos ejemplos, en donde desde la interfaz web se tienen las fotografías de las personas que ingresaron y salieron.

## 4. Conclusiones

En este prototipo de visión artificial se realizaron distintas pruebas en cuanto al *hardware* a utilizar (cámaras 2D y RGB-D) para determinar la mejor opción a implementar en el contador de personas.

Las primeras pruebas que se realizaron con 2D presentaban una complejidad mayor al procesamiento de imágenes debido a factores que son de difícil control en un ambiente natural, como lo fueron la iluminación, los distintos objetos que se percibían en el campo de visión y además el comportamiento de las personas al pasar por una zona que de conteo bidireccional.

Estos retos pusieron a prueba la tecnología básica de captura de imágenes en 2D, lo cual se trabajó modificando el algoritmo de conteo aumentando de 2 a 3 las regiones de interés a procesar, también aumentando la resolución del video para obtener con mejor detalle los objetos en la zona de visión y finalmente trabajando con distintas librerías de procesamiento de imágenes, llegando a un punto aceptable en el resultado final, sin embargo la tecnología de las cámaras 2D aún dejó un amplio margen de error.

En las siguientes etapas de trabajo, al contar con cámaras 3D se migró la idea para intentar mejorarla con esta tecnología y en efecto obtuvimos mejores resultados, uno de los factores que mejoró la calidad del conteo fue el sensor de profundidad con el que cuentan las cámaras 3D, esto fue posible agregando la funcionalidad de segmentar la imagen con una profundidad deseada y permitiendo procesar las imágenes con los objetos que solamente sobrepasaban este umbral ya definido, y esta característica aportó muchísimo a la mejora del proyecto de visión artificial.

Además, esta última característica permitió tres ventajas sobre la tecnología 2D; en primer lugar, acotar el conteo de personas que cumplen la restricción de cumplir con cierta altura (1 metro), segundo optimizar los recursos de hardware para el procesamiento de imágenes ya que la imagen solo contenía un objeto a un metro y como tercera ventaja la mejora en la calidad del conteo en tiempo real.

Finalmente se logró el objetivo de contar personas con un medio no invasivo y en tiempo real con ayuda de la visión artificial. Este prototipo está dirigido a mejorar la calidad y condiciones de trabajo en recintos. Uno de los objetivos primordiales de este proyecto es que se pueda llevar un conteo de personas en cualquier recinto y poder identificar a estas personas para que el usuario o empresa de este sistema pueda manejar situaciones de control o riesgo.

## 5. Bibliografía